**Rukshan Alexander**[(1)], **Prashanthi Rukshan**[(2)], and **Sinnathamby Mahesan**[(3)]

# Natural Language Web Interface for Database (NLWIDB)

**(1)** Faculty of Business Studies, Vavuniya Campus, University of Jaffna, Park Road, Vavuniya, Sri Lanka, rukshan@mail.vau.jfn.ac.lk

**(2)** Sri Lankan Telecom, Colombo Head Office, Sri Lanka, prashanthi_rajadurai@yahoo.com

**(3)** Department of Computer Science, University of Jaffna, Sri Lanka, mahesans@jfn.ac.lk

**Abstract** It is a long term desire of the computer users to minimize the communication gap between the computer and a human. On the other hand, almost all ICT applications store information in to databases and retrieve from them. Retrieving information from the database requires knowledge of technical languages such as Structured Query Language. However majority of the computer users who interact with the databases do not have a technical background and are intimidated by the idea of using languages such as SQL. For above reasons, a Natural Language Web Interface for Database (NLWIDB) has been developed. The NLWIDB allows the user to query the database in a language more like English, through a convenient interface over the Internet.

**Keywords** Natural Language Processing, Natural Language Web Interface

## Introduction

Natural Language Processing (NLP) is interdisciplinary by nature. Linguistics and artificial intelligence can be combined to develop computer programs. It helps to coordinate activities of understanding or producing texts in a natural language. Database Natural Language Processing is an important success in NLP. Asking questions in natural language to get answers from databases is a very convenient and easy method of data access (Androutsopoulos et al., 1995; Siasar djahantighi et al., 2008). The common man does not understand complicated database query languages such as SQL (Structured Query Language) (Siasar djahantighi et al., 2008). Database NLP gives the real-world benefits that can come from these database systems on the one hand, and it works very well in a single-database domain.

Natural Language Interfaces (NLIs) have undergone considerable development since the 70s, but only with moderate success (Androutsopoulos et al., 1995; Thompson et al., 2005) and interest in the topic has consequently decreased since the 90s. However, the necessity for robust and applicable Natural Language Web Interface has become more acute in recent years as the amount of information on the Internet has grown steadily and immensely, and more people from an ever wider population now access data stored in a variety of formal repositories through web browsers (Katz et al., 2002; Minock, 2005). As such, we have begun to address the task of building a Natural Language Web Interface for database.

## Related Work

Databases usually deal with bounded domains such that ambiguity problem in natural language can be resolved successfully (Gauri, 2010). Here are some examples of database NLP systems:

- LUNAR (Woods, 1973) was involved in a system that answered questions about rock samples brought back from the moon. Two databases were used, the chemical analyses and the literature references. The program used an Augmented Transition Network (ATN) parser and Woods' Procedural Semantics. The system was informally demonstrated at the Second Annual Lunar Science Conference in 1971 (Huangi, 2008).

- LIFER/LADDER was one of the first good database language processing systems. It was designed as a natural language interface to a database of information about US Navy ships. This system, as described in by Hendrix (1978), used a semantic grammar to parse questions and query a distributed database. The LIFER/LADDER system could only support simple one-table queries or multiple table queries with easy join conditions (Hendrix, 1978).

- English Wizard is another successful natural language query tool for relational database. It is one of the leading software products that translate ordinary English database requests into Structured Query Language (SQL), and then return the results to the client. English Wizard enables most database reporting tools and client/server applications to understand everyday English requests for information, and also provides graphical user interface (Karande, and Patil, 2009).

## System Description

Natural Language Processing can further facilitate human computer interaction through computer program interface (Rao et al., 2010). Our goal is to produce a Natural Language Web Interface for Database (NLWIDB) to minimize the communication gap between the computer and a human. Furthermore, the aim of NLWIDB system is to facilitate the user to communicate with the computer in a natural way over the Internet. A natural language web interface for accessing information is domain-independent and easy to use without training. The necessity of building the NLWIDB system being able to support users expressing their searching by natural language queries is very important and opens the researching direction with many potential. It combines the traditional methods of information retrieval and the researching of questioning and answering.

A brief description of the NLWIDB system is as follows: consider a database called UNIVERSITY which has been created using MySQL for a University. Within the UNIVERSITY database we create several tables which are properly normalized. Now if the user in global wishes to access the data from the table in the database, he/she has to be technically proficient in the SQL language to make a query for the UNIVERSITY database. Our system eliminates this part and enables the end user to access the tables in his/her own language. Let us take an example: Suppose we want to view information such as year of establishment of department, and code of department which department name equals "Department of Economics and Management" from the department table of the UNIVERSITY database, we use the following SQL statement (query): `SELECT year-of-establishment-of-department, code-of-department FROM department WHERE department-name ='Department of Economics and Management'`. But a person, who doesn't know MySQL database syntax, will not be able to access the UNIVERSITY database unless he/she knows the SQL syntax of firing a query to the database. But using NLP, this task of accessing the database will be much simpler. So the above query will be rewritten using NLP as a question in the web user interface as: `What is the year of establishment of the department and code of the department which department name equals "Department of Economics and Management"?`

Both the SQL statement and NLP statement to access the department table in the UNIVERSITY database would result in the same output by making query not in an SQL like query language, but simply in English like natural language.

**Scope of the System**

The scope of the proposed system as follows,

- To work with any Relational Database Management System (RDBMS) one should know the syntax of the commands of that particular database software (MySQL). A UNIVERSITY database (Fig1) has been created to test the system.
- The interface language is chosen to be English for accommodating wider users.
- Input from the user is taken in the form of questions (like what, who, where).
- All the values in the input natural language statement have to be in double quotes which yield to identify the values from the user input statement.
- A limited data dictionary (sample are given in Tab1) is used where all possible words related to a particular system will be included. The data dictionary of the system must be regularly updated with words that are specific to the particular system.
- Split the question string in to tokens and give order number to each token identified.
- To remove excessive words from the user input statement. Escape words have been considered (sample are given in Tab2) which must be regularly updated with words that are specific to the particular system.
- Rules (Tab3, Tab4, Tab5, Tab6, and Tab7) have been created to perform NLP.
- To identify all SQL elements involved in such as tables, attributes, aggregate, interval, values from the user input statement by using rules.
- To develop an SQL Template. The syntax of the SQL template string for construct an SQL Template as follows:
  [number of ATTRIBUTE, number of TABLE, number of AND, number of AGGREGATE, number of INTERVAL, number of VALUE].
- To construct an SQL query using SQL elements. An algorithm (Fig2) has been developed with the use of Tab8 and Tab9.
- Ambiguity among the words will be taken care of while processing the natural language. For such, the Tab8 has been created to store all attributes belonging to each table.

**System Architecture**

The NLWIDB is concerned with translating user's query from English into its corresponding SQL query to retrieve the data from the relational database via the Internet. The result will then be displayed to the user. This NLWIDB system specially develops for English language as an initial step and other languages will be considered later. An algorithm has been developed efficiently to map a natural language question, entered in English, to convert an SQL statement for producing suitable answers. The algorithm has been implemented with PHP, Apache, MySQL and tested successfully. The NLWIDB system architecture is given in Fig3, which depicts the layout of the process included in converting user question in NL query into a syntactical SQL query to be fired on the RDBMS and getting answers from the database.

The NLWIDB system includes the following modules:
- Graphical User interface: It allows the user to enter the question in a natural language.
- Word Check: It checks all the words in the user question against the data dictionary for its existence.
- Tokenisation: It splits the question string into tokens and gives an order number to each token identified.
- Excess Word Remover: It removes excessive words from the question string.
- Mapping rules: It maps the rules with user input question statement.
- SQL Elements identifier: It identifies all SQL elements involved in the user input statement by using rules and construct the SQL Template String.

- Mapping SQL Template: It identifies correct SQL Template String algorithm for generating SQL query.
- SQL Query Generation: It constructs a query in SQL.
- Run the SQL Query: It gives the SQL query generated to the back-end database.
- Data Collection: This module collects the output of the SQL statement and places it in the user interface screen as a result.

Table 1 List of Data Dictionary

| a | an | and | are | available | code |
|---|---|---|---|---|---|
| department | equal | equals | establishment | exactly | faculty |
| for | Is | Name | of | student | the |
| what | which | which | who | whose | year |

Table 2 List of Escape Words

| a | an | are | available | campus | for |
|---|---|---|---|---|---|
| is | the | what | which | who | whose |

Table 3 Rules representing several different tables in NLWIDB system

| Rule | Rule Symbol | Rule Description |
|---|---|---|
| course | table_course | rule to represent table course |
| courses | table_course | rule to represent table course |
| departments | table_department | rule to represent table department |
| department | table_department | rule to represent table department |
| faculty | table_faculty | rule to represent table faculty |
| faculties | table_faculty | rule to represent table faculty |
| follows | table_follows | rule to represent table follows |
| follow | table_follows | rule to represent table follows |
| student | table_student | rule to represent table student |
| students | table_student | rule to represent table student |

Table 5 Rules indicating several different ways to represent an 'and' or 'as well as' concept

| Rule | Rule Symbol | Rule Description |
|---|---|---|
| and | and_s | rule to represent an and |
| as well as | and_s | rule to represent an and |
| and also | and_s | rule to represent an and |
| also | and_s | rule to represent an and |
| in addition | and_s | rule to represent an and |
| followed by | and_s | rule to represent an and |
| next to | and_s | rule to represent an and |
| with | and_s | rule to represent an and |
| along with | and_s | rule to represent an and |

Table 6 Rules for the aggregate function MAX

| Rule | Rule Symbol | Rule Description |
|---|---|---|
| most | aggregate_max | rule for aggregate function max() |
| maximum | aggregate_max | rule for aggregate function max() |
| highest | aggregate_max | rule for aggregate function max() |
| biggest | aggregate_max | rule for aggregate function max() |
| maximum number of | aggregate_max | rule for aggregate function max() |
| highest number of | aggregate_max | rule for aggregate function max() |
| biggest number of | aggregate_max | rule for aggregate function max() |

Table 7 Rules indicating several different ways to represent interval 'equal' concept

| Rule | Rule Symbol | Rule Description |
|---|---|---|
| equal | interval_= | rule for interval equal |
| exactly | interval_= | rule for interval equal |
| equals | interval_= | rule for interval equal |

Table 4 Rules representing different attributes of department tables in NLWIDB system

| Rule | Rule Symbol | Rule Description |
|---|---|---|
| code of department | attribute_department_code | rule to represent attribute department code |
| department code | attribute_department_code | rule to represent attribute department code |
| department name | attribute_department_name | rule to represent attribute department name |
| department names | attribute_department_name | rule to represent attribute department name |
| name of department | attribute_department_name | rule to represent attribute department name |
| names of departments | attribute_department_name | rule to represent attribute department name |
| year of establishment of department | attribute_department_year_of_establishment | rule to represent attribute department year of establishment |
| year of establishment of departments | attribute_department_year_of_establishment | rule to represent attribute department year of establishment |

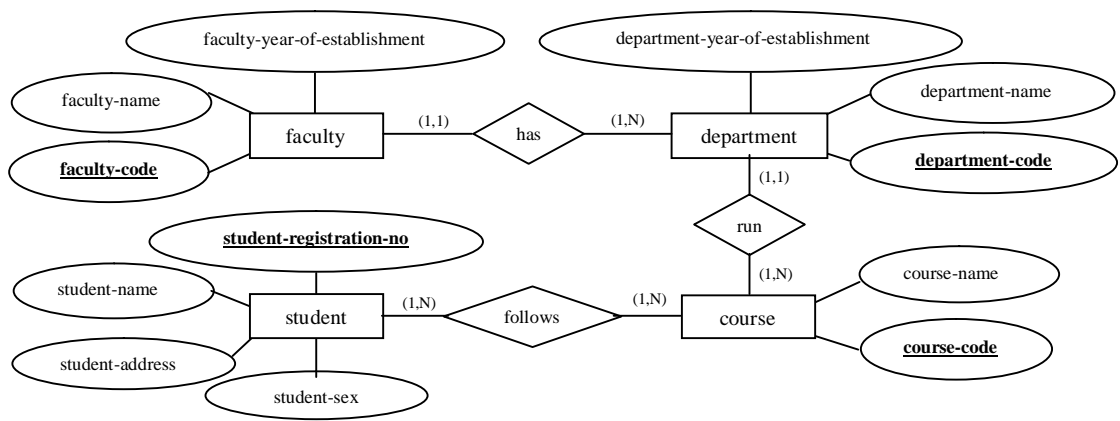

Figure 1 The UNIVERSITY database

```
Begin
    Store the SQL template string to the variable U
    If (the value of variable U equals '020000') then
        Store the name of the identified tables in array V
        Do
            Store the value of variable V to T
            Get the default attribute for the table name T and store it to the variable A
            Construct a SQL query as "SELECT DISTINCT A FROM T" and store it to variable W
        While for each table in array V
    End If
End
```

Figure 2 Algorithm to construct the SQL query for the *SQL Template String:* 020000

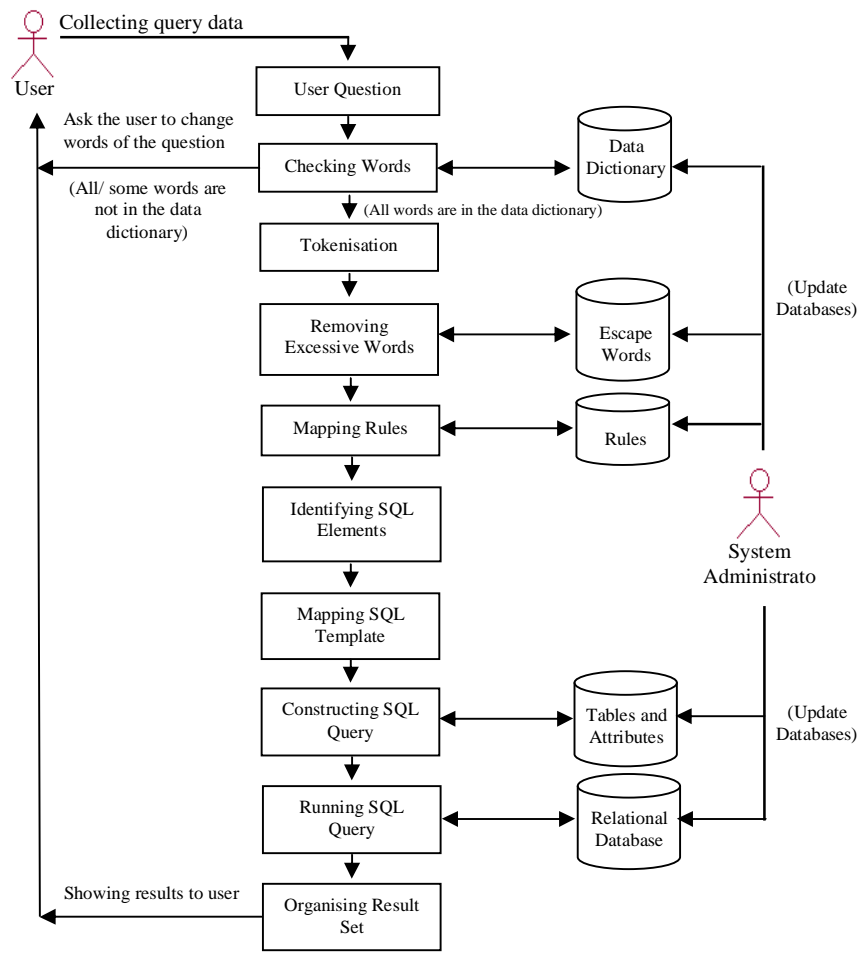

Figure 3 Detail Architecture of NLWIDB System

Table 8 Database tables and attributes

| Table Name | Attribute Name | Key |
|---|---|---|
| department | department-name | nil |
| department | department-code | primary |
| department | department-year-of-establishment | nil |
| department | faculty-code | foreign |
| faculty | faculty-code | primary |
| faculty | faculty-name | nil |
| faculty | faculty-year-of-establishment | nil |

Table 9 Database default attributes in tables

| Table | Default Attribute |
|---|---|
| department | department-name' |
| faculty | faculty-name |
| course | course-name |
| student | student-registration-no' |

**Algorithm used in NLWIDB System**

The following steps depict the algorithm for the NLWIDB system

- Check for an input value and if exists remove the value from the question string.
- Check for all words in question string which exists in data dictionary.
- Tokenization (scanning)
    - Split the question string in tokens
    - Give order number to each token identified
- Removing excessive words in question string.
- Mapping rules by removing last words from question string. If mapped with a rule, remove the mapped string from the question string and make the rest as the question string. Map the rule until you have question string becomes null.
- Store the number of identified SQL elements in arrays.
- Construct an SQL template string by considering available number of SQL elements.
- Mapping SQL template string with SQL template.
- Construct SQL query using SQL elements.
- Get result set.
- Organise the result.
- Display the result to user.

## Results and Discussion

Consider following relations (tables in database) department (Tab10) and faculty (Tab11) with data.

The Fig5 indicates that the results which were computed by the NLWIDB system when the user gives the input statement as: What is the year of establishment of department and code of department of which department name equals "Department of Economics and Management"? at the user input text of the NLWIDB (Fig4). Furthermore, the Tab12 describes the list of variety of questions which our system deals:

Table 10 The department table with data

| department-code | faculty-code | department-name | department -year-of-establishment |
|---|---|---|---|
| DACF | FBS | Department of Accountancy and Finance | 1997 |
| DECM | FBS | Department of Economics and Management | 1997 |
| DBIS | FAS | Department of Bio Science | 1997 |
| DPHS | FAS | Department of Physical Science | 1997 |

Table 11 The faculty table with data

| faculty_code | faculty-name | faculty-year-of-establishment |
|---|---|---|
| FAS | Faculty of Applied Sciences | 1997 |
| FBS | Faculty of Business Studies | 1997 |

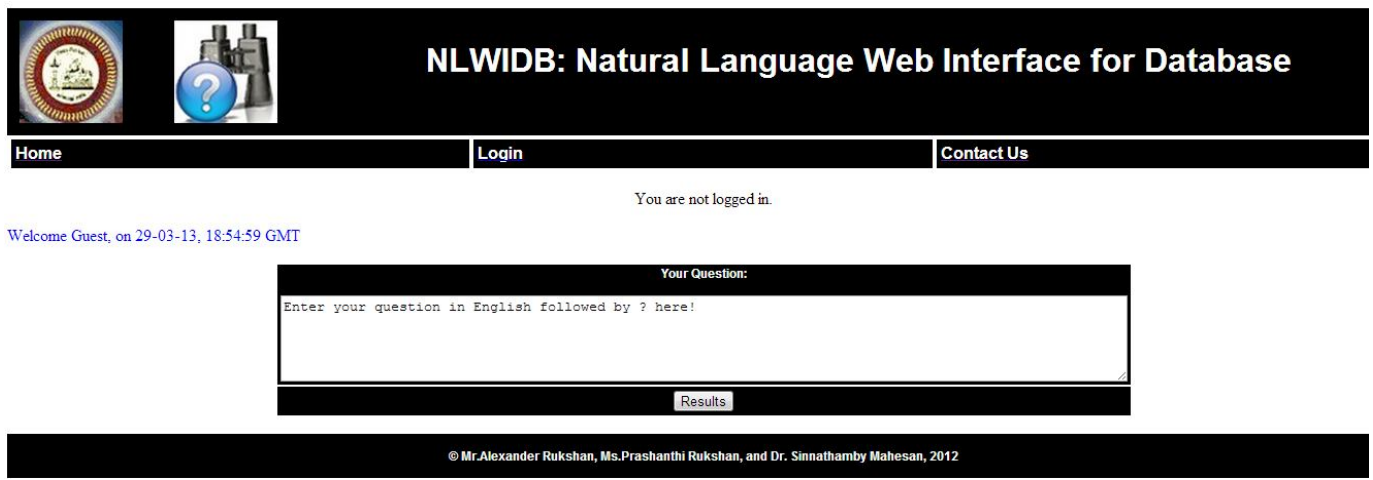

Figure 4 Graphical User Interface of NLWIDB System (User input screen)

```
User Input Natural Language Statement as Question:
************
What is the year of establishment of department and code of department which department name
equals "Department of Economics and Management"?
---------------------------------------------------------------
Checking Words:
*************
The user question after removing value: What is the year of establishment of department and code of department which department name
equals
All words in the user question are exists in the Data Dictionary.
---------------------------------------------------------------
Tokenisation:
**********
The token [0]: What
The token [1]: is
The token [2]: the
The token [3]: year
The token [4]: of
The token [5]: establishment
The token [6]: of
The token [7]: department
The token [8]: and
The token [9]: code
The token [10]: of
The token [11]: department
The token [12]: which
The token [13]: department
The token [14]: name
The token [15]: equals
The token [16]: "Department
The token [17]: of
The token [18]: Economics
The token [19]: and
The token [20]: Management"
---------------------------------------------------------------
Removing Excessive Words:
***********************
The question string after removing excessive words for mapping rule:
year of establishment of department and code of department department name equals "Department of Economics and Management"
---------------------------------------------------------------
Mapping Rules:
*************
The string for rule mapping: year of establishment of department and code of department department name equals "Department of
Economics and Management"
The string after removing value: year of establishment of department and code of department department name equals
The string after removing last word is "year of establishment of department and code of department department name"
The string after removing last word is "year of establishment of department and code of department department"
The string after removing last word is "year of establishment of department and code of department"
The string after removing last word is "year of establishment of department and code of"
The string after removing last word is "year of establishment of department and code"
The string after removing last word is "year of establishment of department and"
The string after removing last word is "year of establishment of department"
 • The string "year of establishment of department" mapped with rule symbol: attribute_department_year_of_establishment
The string after removing last word is "and code of department department name"
The string after removing last word is "and code of department department"
The string after removing last word is "and code of department"
The string after removing last word is "and code of"
The string after removing last word is "and code"
The string after removing last word is "and"
 • The string "and" mapped with rule symbol: and_s
The string after removing last word is "code of department department name"
The string after removing last word is "code of department department"
The string after removing last word is "code of department"
```

```
• The string "code of department" mapped with rule symbol: attribute_department_code
The string after removing last word is "department name"
• The string "department name" mapped with rule symbol: attribute_department_name
• The string "equals" mapped with rule symbol: interval_=
----------------------------------------------
Identifying SQL Elements:
*******************
Attribute: department-year-of-establishment
Attribute: department-code
Attribute: department-name
And: s
Interval: =
Value: Department of Economics and Management
----------------------------------------------
Mapping SQL Template:
*******************
The string code for finding SQL Template [attribute, table, and, aggregate, interval, value]: m01011
Finding SQL template for constructing SQL query...
Found SQL template.
----------------------------------------------
Constructing SQL Query:
***********************
The SQL query: select department-year-of-establishment,department-code,department-name from department where department-name = 'Department of Economics and Management'
----------------------------------------------
Running SQL Query:
*******************
The SQL query is passed to RDBMS for getting results
----------------------------------------------
Organising Result Set:
*******************
```

| Department year of establishment | Department code | Department name |
|---|---|---|
| 1997 | DECM | Department of Economics and Management |

Figure 5 Results computed by the NLWIDB

Table 12 list of variety of questions which the NLWIDB deals

| Q No | User Input Question | Identified SQL Template | NLWIDB Output |
|---|---|---|---|
| 1 | What are the available names of departments? | 100000 | **Department Name**<br>Department of Accountancy and Finance<br>Department of Bio Science<br>Department of Economics and Management<br>Department of Physical Science |
| 2 | What are the available departments? | 010000 | **Department Name**<br>Department of Accountancy and Finance<br>Department of Bio Science<br>Department of Economics and Management<br>Department of Physical Science |

| 3 | What are the available departments and faculties? | 0m1000 | Department Name |||
|---|---|---|---|---|---|
| | | | Department of Accountancy and Finance |||
| | | | Department of Bio Science |||
| | | | Department of Economics and Management |||
| | | | Department of Physical Science |||
| | | | Faculty Name |||
| | | | Faculty of Applied Sciences |||
| | | | Faculty of Business Studies |||
| 4 | What is the maximum year of establishment of departments? | 100100 | Department Year of Establishment |||
| | | | 1997 |||
| 5 | What are the department names which year of establishment of department equals "1997"? | 110011 | Department Name | Department Year of Establishment ||
| | | | Department of Accountancy and Finance | 1997 ||
| | | | Department of Bio Science | 1997 ||
| | | | Department of Economics and Management | 1997 ||
| | | | Department of Physical Science | 1997 ||

## Conclusion

Natural Language Processing can bring powerful enhancements to virtually any web interface, because human language is so natural and easy to use for humans. The Natural Language Web Interface for Database (NLWIDB) is no exception because we presented the NLWIDB system which converts a wide range of text queries (English questions) into formal ones that can then be executed against a database by employing robust language processing techniques and methods. This research shows that NLWIDB provides a convenient as well as reliable means of querying access, hence, a realistic potential for bridging the gap between computer and the casual end users. While NLP is a relatively recent area of research and application, as compared to other information technology approaches, there have been sufficient successes to date that suggest that NLP-based information access technologies will continue to be a major area of research and development in information systems now and far into the future. Furthermore, we could bypass a prohibitive formal query language that can only be reasonably managed by experts while still offering rich tools for the composition of complex queries by casual users.